\documentclass{article} % DO NOT CHANGE THIS
\usepackage{aaai23}  % DO NOT CHANGE THIS
\usepackage{times}  % DO NOT CHANGE THIS
\usepackage{helvet}  % DO NOT CHANGE THIS
\usepackage{courier}  % DO NOT CHANGE THIS
\usepackage[hyphens]{url}  % DO NOT CHANGE THIS
\usepackage{graphicx} % DO NOT CHANGE THIS
\usepackage{natbib}  % DO NOT CHANGE THIS AND DO NOT ADD ANY OPTIONS TO IT
\usepackage{caption} % DO NOT CHANGE THIS AND DO NOT ADD ANY OPTIONS TO IT
\usepackage[ruled, linesnumbered]{algorithm2e}
\usepackage{algorithmic}

\usepackage{amsfonts}
\usepackage{amsmath}
\usepackage{pifont}
\usepackage{booktabs}
\newcommand{\cmark}{\ding{51}}%
\newcommand{\xmark}{\ding{55}}%
\usepackage[dvipsnames]{xcolor}
\usepackage{newfloat}
\usepackage{listings}
\DeclareCaptionStyle{ruled}{labelfont=normalfont,labelsep=colon,strut=off} % DO NOT CHANGE THIS
\nocopyright %-- Your paper will not be published if you use this command
\title{SELF-VS: Self-supervised Encoding Learning For Video Summarization}
\author {
    Hojjat Mokhtarabadi\equalcontrib\textsuperscript{\rm 1},
    Kave Bahraman\equalcontrib\textsuperscript{\rm 1},
    Mehrdad HosseinZadeh\textsuperscript{\rm 2},
    Mahdi Eftekhari\textsuperscript{\rm 1} 
}

\affiliations{
    \textsuperscript{\rm 1}Department of Computer Engineering, Shahid Bahonar University of Kerman, Iran\\
    \textsuperscript{\rm 2} University of Manitoba\\

    \{hojjat.mokhtarabadi, sbahraman, m.eftekhari\}@eng.uk.ac.ir, mehrdad@cs.umanitoba.ca
}

\usepackage{bibentry}
\begin{document}

\maketitle

\begin{abstract}
Despite its wide range of applications, video summarization is still held back by the scarcity of extensive datasets, largely due to the labor-intensive and costly nature of frame-level annotations. As a result, existing video summarization methods are prone to overfitting. To mitigate this challenge, we propose a novel self-supervised video representation learning method using knowledge distillation to pre-train a transformer encoder. Our method matches its semantic video representation, which is constructed with respect to frame importance scores, to a representation derived from a CNN trained on video classification. Empirical evaluations on correlation-based metrics, such as Kendall's $\tau$ and Spearman's $\rho$ demonstrate the superiority of our approach compared to existing state-of-the-art methods in assigning relative scores to the input frames.
\end{abstract}

\section{Introduction}
The explosion of video content on the internet has led to a growing interest in the field of video summarization. With more than 500 hours of video content being uploaded to YouTube every minute, it is becoming increasingly difficult for users to sift through and find relevant information. The goal of video summarization is to condense long videos into shorter clips while preserving the key semantic information. This is particularly useful for applications such as video search and retrieval, video surveillance, and video-based educational systems. Generally speaking, video summarization techniques can be classified into two categories: \begin{enumerate}\item keyframe-based \item key-shot-based\end{enumerate} The major distinctive factor between these two settings is the granularity level that is considered in a summary generation; the former picks important frames to form a static summary (i.e., storyboard), while the latter first segments a video into shots and then selects the most informative shots to form a dynamic summary (i.e., video skim). This work focuses on the keyframe-based setting.

One of the challenges in video summarization is the lack of large, labeled datasets for training neural networks. Human annotation of videos is a costly and tedious task, and existing datasets such as TVSum\cite{song2015TVSum} and SumMe\cite{gygli2014creating} are not large enough for training modern neural networks such as transformers\cite{vaswani2017attention}. This is a significant barrier to the development of effective video summarization models.

Recently, \cite{Li2021WeaklySD} proposed a semantically meaningful reward as the similarity between video and generated summary representations. One of the crucial aspects of video summarization is ensuring that the generated summary preserves the semantics of the source video. Motivated by this concept, our approach applies a novel form of self-supervised learning with knowledge distillation which not only allows for efficient utilization of unlabelled data but to generate a summary that maintains semantic similarity to the source video. Our approach is different in a sense that we explicitly try to learn frame scores during pre-training.

\begin{figure}
    \centering
    \includegraphics[width=0.45 \textwidth]{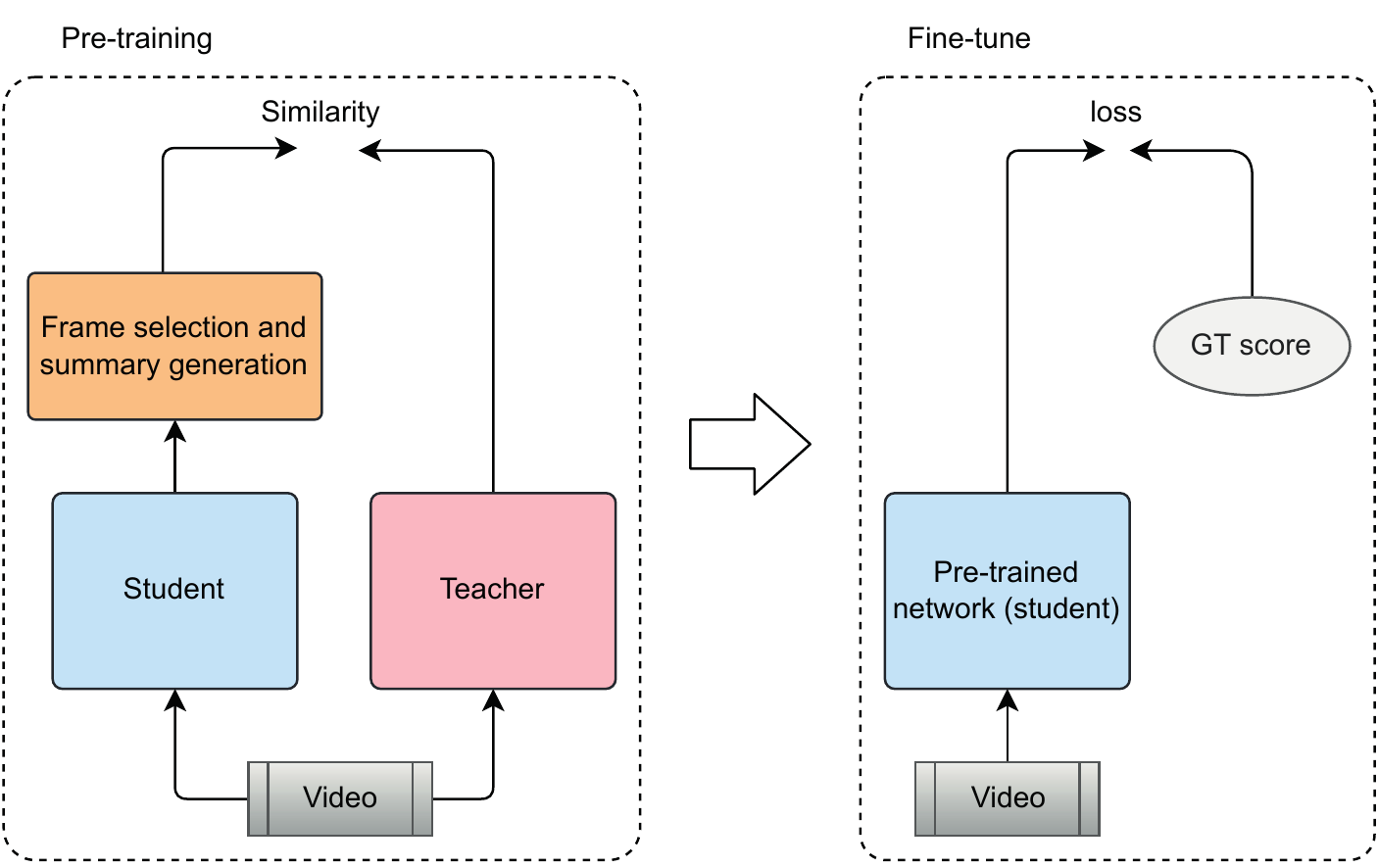}
    \caption{
    The main objective of our proposed framework is to maximize the similarity between video and summary representations through a teacher-student pre-training process, where the teacher provides a supervisory signal for the student. In the fine-tuning phase, only the pre-trained student is used and teacher is discarded.}
    % The students network is pre-trained with supervision signal from teacher. This network is then fine-tuned on ground truth scores.} 
    \label{fig:5}
\end{figure}

% kave was here
% Our goal is to maximize the similarity between the video representation and the generated summary representation. Thereby, at

More specifically, our proposed method employs a teacher-student scheme in which the teacher provides a global representation (self-supervisory signal) of the video, and the student network selects a subset of frames that closely match to the teacher's representation. The selection procedure is done by assigning a relative score to each frame and our primary choice for teacher and student networks are transformer encoder and a pre-trained 3D CNN respectively.

As discussed above the benefit of using a self-supervised setting is that the supervision signal is obtained from the data itself, often by leveraging the underlying structure in the data. This enables the model to learn useful representations of video without the need for expensive and time-consuming manual annotation. Despite the simplicity of the design, our method attains state-of-the-art performance in regards to correlation metrics, namely Kendall's $\tau$ and Spearman's $\rho$.

The key contributions of this research are summarized as:
\begin{itemize}
    \item We address the lack of large-scale annotated dataset by leveraging a self-supervised training scheme. 
    \item We introduce a novel integeration of teacher-student and self-supervised learning frameworks with the hybrid training scheme. 
    \item Finally, we show that the proposed method achieves state-of-the-art performance on benchmark datasets through extensive experiments.
\end{itemize}
% intro ends

\section{Related work}
This section briefly overviews two main categories of video summarization as well as self-supervised learning framework.

\subsection{Supervised video summarization} 
These methods aim to optimize the summary generation process using human-provided annotations. Early works in this area, such as \cite{zhang2016video} exploited LSTM networks to model variable-range temporal dependencies between video frames. Subsequently, \cite{Zhao2017HierarchicalRN} addressed the limitations of vanilla LSTM models by proposing a hierarchical LSTM architecture. To further improve the capture of temporal dependencies, \cite{ji2019video} introduced an attention mechanism. However, global attention models were found to be inadequate in capturing local dependencies, leading to more recent approaches such as \cite{Pan2022ExploringGD} that propose a combination of local and global attention mechanisms. Despite the progress made in this area, the need for human annotations for training remains a significant limitation, as it can be costly and infeasible in certain scenarios such as military operations.

\subsection{Unsupervised video summarization} As opposed to supervised methods these techniques attempt to generate summaries without the need for human annotations. Techniques such as \cite{apostolidis2020unsupervised, fu2019attentive, zhang2019dtr, mahasseni2017unsupervised} employ GANs to distinguish between generated summaries and the original videos. Other methods such as \cite{zhou2018deep, li2021exploring} optimize summary generation using human-designed criteria such as frame-level diversity and representativeness. Recent works, such as \cite{li2021weakly}, also incorporates semantic meaningful rewards in addition to diversity and representativeness for improved results.

Despite the progress made in these areas, a common limitation of existing video summarization methods is their reliance on small-scale datasets for training neural networks.

\subsection{Self-Supervised Leanring}
Supervised learning has mostly been a dominant paradigm of learning, however as the AI field is progressing, gathering massive amounts of labeled data can become a bottleneck. In the face of this problem, Self-supervised learning (SSL) has emerged as a promising alternative to traditional supervised learning in recent years. This approach utilizes the data itself as the supervisory signal, allowing for the learning of universal representations that can be beneficial for downstream tasks

Self-supervised learning has had significant impact on natural language processing (NLP) models, such as BERT \cite{devlin-etal-2019-bert} and RoBERTa \cite{liu2019roberta}, which have been pre-trained in a self-supervised manner and then fine-tuned on downstream tasks such as text classification. Yet in other domains like computer vision, it's been more challenging to follow this paradigm. Early works in this area, such as \cite{Noroozi2016UnsupervisedLO, Gidaris2018UnsupervisedRL, Larsson2016LearningRF}, focused on changing the data and then asking the network to recognize those changes. More recent works, such as \cite{chen2020simple, He2020MomentumCF, grill2020bootstrap, caron2020unsupervised}, have proposed contrastive and non-contrastive SSL approaches for image data.  In the former, the main goal is to augment images into different views and then ask the network which augments originate from the same image whereas in the latter the goal is to approximate views of the same image independent from other images. This work follows a non-contrastive self-supervised learning approach for video summarization.
% related ends

\section{Background}
\subsection{Transformer}
The transformer architecture \cite{vaswani2017attention} is widely regarded as one of the breakthroughs in the field of deep learning. This architecture is composed of two main modules, the encoder and the decoder, which are placed side by side. In this work, we utilize the encoder module, which consists of multiple stacked layers. Each layer is divided into two sub-layers: a self-attention (SA) sub-layer and a position-wise feed-forward network (FFN) sub-layer.

\subsubsection*{Self attention (SA)}
Given an input sequence $\{x_i\}_{i=0}^T$ with the length T and $x_i \in \mathbf{R}^{d}$, SA is responsible for mapping input vectors to new representations based on their similarity to other vectors in the sequence. To be more precise, every vector in the sequence is transformed to a query, key and value vector via linear transformations represented by the matrices $W_q$, $W_k$, $W_v \in \mathbf{R}^{d \times a}$ respectively, and $a$ also is the attention space dimension. The new representation is then calculated as a weighted sum of the value vectors, with the weights being determined by the normalized compatibility score between the paired key and query vectors. This is formulated as below:
\begin{equation}
    Attention(Q, K, V) = softmax(\frac{Q K^T}{\sqrt{d}}) V
\end{equation}
where $Q$, $K$, $V \in \mathbf{R}^{T \times a}$  are all query, key, and value vectors bundled together. Note that the compatibility function is scaled dot-product and the normalizing is a softmax between scores for each query. 

\subsubsection*{Position-Wised Feed forward network (FFN)}
In addition to the self-attention sub-layer, a feed-forward network is applied to each vector independently. This sub-layer consists of two linear layers with an activation function in between:
\begin{equation}
    FFN(x) = Activation(x W_1 + b_1)W_2 + b_2
\end{equation}
Main choice of activation function is ReLU\cite{agarap2018deep}, however in \cite{dosovitskiy2021an} authors show that is beneficial to use GeLU\cite{Hendrycks2016GaussianEL}.

\subsection{Knowledge Distillation}
Knowledge distillation, first proposed by \cite{hinton2015distilling} is a technique that allows for the transfer of knowledge from a pre-trained model, known as the teacher, to a smaller model, known as the student. The basic idea behind knowledge distillation is that the student model will be able to improve its performance by mimicking the behavior of the teacher model. This method is mostly utilized for model compression and learning from multiple teacher. However, recently in \cite{touvron2021training} authors demonstrated that distillation training can not only encapsulate knowledge in smaller models but also aid in situations where there is a shortage of proper amounts of data.  This is known as "soft-distillation" and is formulated by the following equation:
\begin{equation}
\begin{split}
    \mathcal{L}_{global} & = (1 -\lambda) \mathcal{L}_{CE}( \psi( Z_s), y)\\ &+ \lambda{\tau}^2 \mathrm{KL}( \psi( Z_s/\tau ), \psi( Z_t/\tau ))
   \label{equ:0}
\end{split}
\end{equation}
where $Z_t$ is the logits of the teacher model and $Z_s$ the logits of the student model. $\tau$ is denoted as the temperature for the distillation, $\lambda$ as a coefficient balancing the Kullback–Leibler divergence loss ($\mathrm{KL}$) and the cross-entropy ($\mathcal{L}_{CE}$) on ground truth labels y, and $\psi$ is the softmax function.

\section{Method}
Deep learning models require large amounts of data to be trained effectively. However, recent studies \cite{zhou2018deep, mahasseni2017unsupervised, jung2019discriminative, apostolidis2020ac} have shown that current video summarization datasets are not sufficient for proper training of fully supervised models.

This problem can be addressed by a hybrid training scheme where a model is first trained on a large-scale annotated (and related) dataset, and then is fine-tuned on a downstream task with a small amount of annotated data. Nevertheless, the need for a large annotated dataset for pre-training still remains. The problem even amplifies in the video domain as annotations for videos are more costly to collect than for images. 

To circumvent this obstacle, we propose a self-supervised learning (SSL) scheme based on a teacher-student framework where publicly available videos can be utilized to train a video representation learning model. The pre-trained model is then fine-tuned on current video summarization datasets. In this section, we detail the proposed model, then further discuss the approach to avoid collapse and finally elaborate on the loss function and fine-tune phase.

\begin{figure*}[t]
    \centering
    \includegraphics[width=1.\textwidth]{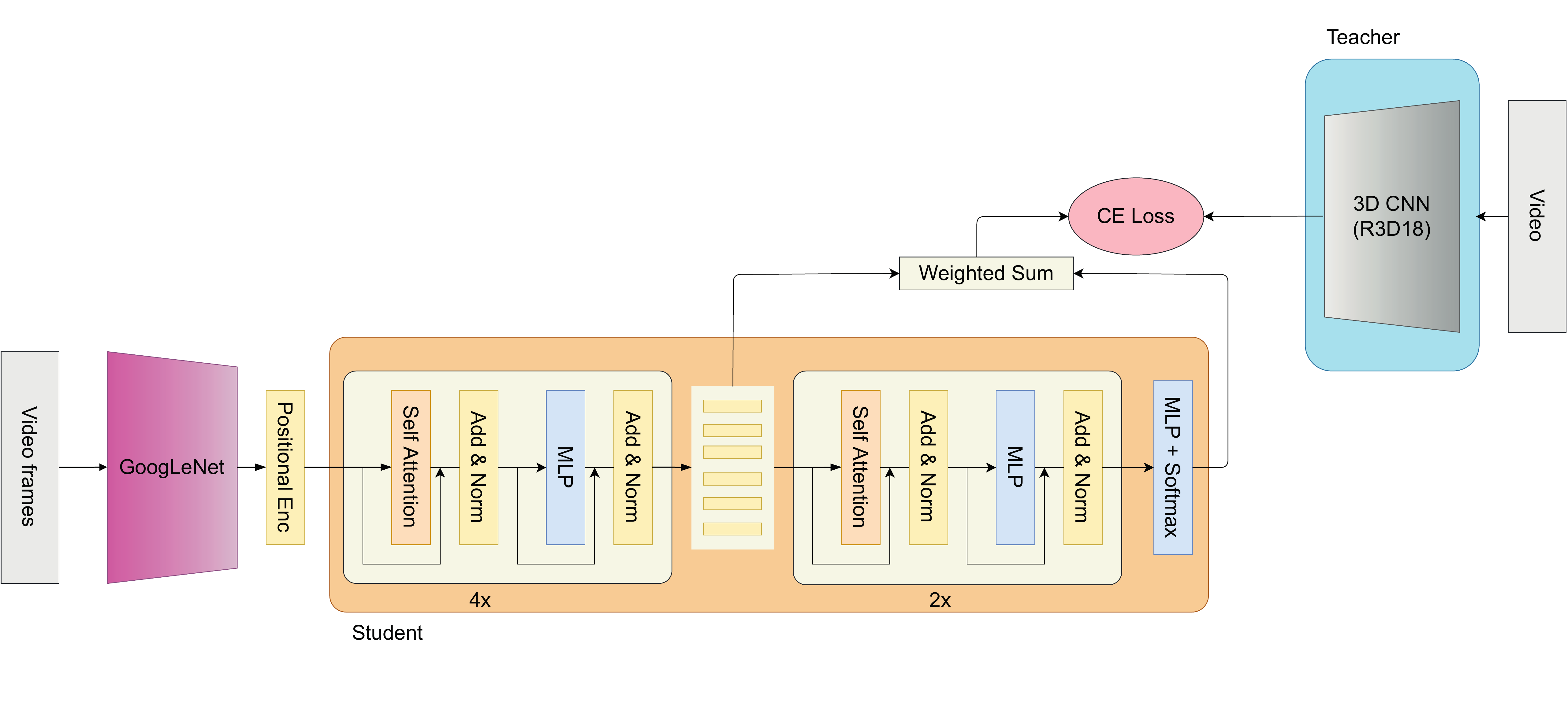}
    \caption{The student network consists of two serial encoders followed by an MLP with softmax activation to assign a relative score to each frame. The teacher network is a simple CNN that takes a video and outputs its representation. Encoder video representations are constructed by a weighted sum between intermediate frame features where weights are the scores assigned to each frame.
} 
    \label{fig:1}
\end{figure*}

\subsection{SSL pre-tranining}

 % define comment color
\definecolor{commentcolor}{RGB}{68, 68, 68}  
\newcommand{\PyComment}[1]{\ttfamily\textcolor{commentcolor}{\# #1}}  % add a "#" before the input text "#1"
\newcommand{\PyCode}[1]{\ttfamily\textcolor{black}{#1}} % \ttfamily is the code font
\newcommand{\PyKey}[1]{\ttfamily\textcolor{black}{#1}} % \ttfamily is the code font

\begin{algorithm}[ht]
    \SetAlgoLined
    \PyComment{T: pre-trained teacher network}\\ 
    \PyComment {S: student network} \\
    \PyComment{$x_t$: video (batch\_size, 3, n\_frames, H, W)} \\
    \PyComment{$x_s$: video frames (batch\_size, n\_frames, frame\_dim)}\\
    \PyKey{for} \PyCode{$x_s$, $x_t$} \PyKey{in} \PyCode{loader:} \\
    \Indp
        \PyComment{w: the weight vector for video frames}\\
        \PyCode{frame\_ft, w = S($x_s$)} \\
        \PyCode{cr = T($x_t$)} \PyComment{CNN representation}
        
        \BlankLine
        \PyComment{weighted sum to construct encoder video representation er} \\
        \PyCode{er = torch.matmul(weights, frame\_ft)} \\
        
        \BlankLine 
        \PyComment{loss} \\
        \PyCode{loss = cross\_entropy(er, cr)} \\ % + (rep\_loss(frame\_ft)} * $\lambda$)\\
        \PyCode{loss.backward()} \\
        \PyComment{optimize only student} \\
        \PyCode{optimizer.step()} \\
    
    \Indm 
    
    \BlankLine
    \PyComment{CE loss}\\
    \PyKey{def} \PyCode{cross\_entropy(x, y):}\\
    \Indp
        \PyCode{x = F.softmax(x, dim=1)}\\
        \PyCode{y = F.softmax(y, dim=1)}\\
        \PyComment{} \\
        \PyCode{loss = -y * torch.log(x)}\\
        \PyKey{return} \PyCode{loss.mean()} \\
    \Indm
    
%    \BlankLine
%    \PyComment{Repelling Loss} \\
%    \PyKey{def} \PyCode{rep\_loss(x):} \\
%    \Indp
%        \PyComment{long code omitted} \\
%        \PyKey{return} \PyCode{sim} \PyComment{between [-1, 1]}\\
%    \Indm

\caption{PyTorch-style pseudocode for SELF-VS}
\label{algo:your-algo}
\end{algorithm}

Our proposed SSL pre-training approach involves the use of two networks: a 3D CNN network trained for video classification as the teacher and a transformer-based encoder as the student. The goal is to leverage the pre-trained 3D CNN to provide a supervisory signal for training the encoder. The intuition behind this approach is that a CNN trained for video classification is capable of capturing the semantics of the entire video. This intuition is particularly beneficial when viewing video summarization as the selection of important frames. By combining frame representations acquired with respect to their relative scores, the student network can be forced to give higher scores to important frames.

In more detail, the video representation is obtained by feeding raw video frames to the teacher network (i.e., the 3D CNN). The student network then seeks to learn this representation by assigning scores to the video frames and using these scores to compute a weighted sum of its own intermediate frame features. In other words, the student network learns both the frame features and frame importance scores simultaneously. This novel SSL scheme enables the student network to perform well on the downstream task of video summarization.

Given a video $v$ with $n$ frames, to acquire the representation from the student network $\phi_v$, intermediate frame features are aggregated with respect to their corresponding scores.\\
More precisely, at first image features are extracted from the GoogLeNet network, and positional embeddings are then concatenated with them, similar to the approach proposed by \cite{vaswani2017attention}. This allows for the inclusion of both visual and positional information in the extracted features. Resulted frame features are then fed to the first encoder which outputs intermediate features $X =\{x_i\}^n\in \mathbb{R}^{n \times d}$.

The intermediate features obtained from the first encoder are subsequently used as input for the second encoder, in conjunction with a position-wise linear transformation $W_s \in \mathbb{R}^{d}$. These scores are then normalized with softmax to get aggregation weights:
\begin{equation}
    W = softmax(X W_s)
\end{equation}
where $W = \{w_i\}^n$ contains the frame-level scores.
These weights are used in a weighted sum between frame representations to get video representation:
\begin{equation}
    \phi_v = \sum_{i=1}^{n} w_i\ x_i
\end{equation}
Because the network learns to assign importance value to frames, this approach can be very effective specifically in video summarization.

%\subsubsection*{Avoiding Collapse}
%Aggregation weights are used in the weighted sum between frames to obtain transformer video representation. This procedure makes it possible for the network to completely disregard its design purpose and assign the same weight to all frames. Then the network can cheat by transforming every frame representation to video representation which leads to uniform collapse.

%However to solve this issue it is sufficient to push frames representation further from each other. To this end we employ a repelling loss \cite{mahasseni2017unsupervised} which aims to minimize the cosine similarity between all frame.
%\begin{equation}
%    \mathcal{L}_{repelling} = \frac{1}{T} \frac{1}{T-1} \sum_{i} \sum_{i^{\prime} \neq i} \frac{e_i^T e_{i^{\prime}}}{||e_i||\ ||e_{i^{\prime}||}}
%\end{equation}
%where $e_i$ is the representation of the i-th frame and $T$ is the number of video frames. 

\subsubsection*{Knowledge distillation loss}
As mentioned before, knowledge distillation is adapted to guide the training process from a pre-trained CNN teacher. In order to align the teacher and student's representations, initially both output vectors that are obtained from teacher and student model are passed through a softmax function to obtain an output distribution. Then the training process is guided by minimizing the cross-entropy between the CNN's and encoder's output distribution. The loss function is formulated as below:
\begin{equation}
    \mathcal{L}_{dis}(a, b) = - \sum_{i=0}^{n} a_i \log b_i
\end{equation}
where $a,b \in \mathbb{R}^{n}$ vectors represent CNN and encoder output distribution, respectively.
Note that this is a special form of the equation \ref{equ:0} where the left term is absent because there are no supervision labels. The right term can be written as following:

\begin{equation}
    KL(a, b) = \sum_{i=0}^{n} a_i \log a_i -  \sum_{i=0}^{n} a_i \log b_i
\end{equation}
where the first term is the entropy of CNN distribution and the second term is the cross-entropy between CNN and the encoder distribution.
% Because CNN weights are frozen the left term is always constant so can be discarded and we only optimize on the right one.
In our case, the left term is discarded because CNN weights are frozen so the left term is always constant and only the right one is optimized. 

%Taking into the account for weight penalization, the final SSL loss function is formulated as below:
%\begin{equation}
%        \mathcal{L}_{SSL} = \mathcal{L}_{dis} + \lambda \mathcal{L}_{repelling}
%\end{equation}
%where $\lambda \in [0,1]$ is the coefficient controlling centering.

\subsection{Fine-tuneing} We follow standard training procedure of fine-tuning the network in a supervised manner. However, instead of randomly initializing the weights, the pre-trained weights obtained from the previous phase are used as the starting point.
To optimize the network, mean squared error (MSE) between ground truth scores and scores obtained from the network is used as the loss function. 
% The goal for the network is to assign a score close to the ground truth for each frame. This can be formulated using a supervised loss as below:
\begin{equation}
    \mathcal{L}_{S} = \frac{1}{M} \sum_{i=0}^{n_j} (s_i - \hat{s_i})^2
\end{equation}

where $n_j$ is the number of frames in the jth video and $M$ is the total number of videos and $s_i, \hat{s_i}$ are the ground truth and the network scores respectively.

\section{Experimental results}
In this section, we provide an insight into the experimental environment and compare our approach to other state-of-the-art methods. Furthermore, we study the effect of pre-training and give qualitative results.

% \subsection{Setup}
\subsubsection{Dataset}
For pre-training, we make use of UCF101\cite{soomro2012ucf101} which is an action recognition dataset that contains about 13k videos with durations ranging from 2 to 10 seconds. Additionally, the fine-tuning and evaluation steps are done on two benchmark datasets, TVSum\cite{song2015TVSum} and SumMe\cite{gygli2014creating}. TVSum consists of 50 YouTube videos from the TRECVid Multimedia Event Detection (MED) \cite{smeaton2006evaluation}. Videos in TVSum vary over 10 different categories where each category shares 5 videos. Each video ranges between 2 to 10 minutes and is annotated by 20 different users. To annotate a video, each user assigns a score between 1-5 to its shots, then these scores are converted to ground truth scores following the procedure in \cite{song2015TVSum}. SumMe is a collection of 25 videos varying in category and camera angle, i.e., first person, third person, and egocentric. Each video in SumMe ranges between 90 to 390 seconds and is annotated by multiple users, therefore multiple ground truth summaries are averaged to form the final ground truth for training.

%\begin{table}
%\centering
%\begin{tabular}{c|c|c}
%\hline
%Dataset & Numer of videos & Type of annotation          \\
%\hline
%SumMe   & 25              & Set of key fragments        \\
%TVSum   & 50              & Fragment-level score        \\
%OVP     & 50              & Multiple sets of key-frames \\
%Youtube & 50              & Multiple sets of key-frames \\
%\hline
%\end{tabular}
%\caption{Widely used video summarization datasets. Mixing all datasets results %in 226 sample dataset which is still insufficient for training deep neural networks.}
%\label{table:1}
%\end{table}

\begin{table}
\centering
\begin{tabular}{ccc}
\toprule
Method  & SumMe & TVSum\\
\midrule
DR-DSN\shortcite{zhou2018deep}      & 42.10  &  58.1\\
SGAN\shortcite{Mahasseni_2017_CVPR}         & 38.7  &  50.8 \\
CLIP-It$_{uns}$\shortcite{NEURIPS2021_7503cfac}  &  52.5  &  63.0\\
SSPVS\shortcite{haopeng2022video}  & 49.8 & 60.1 \\
\midrule
dppLSTM\shortcite{zhang2016video}     & 38.60 & 54.70\\
S-FCN\shortcite{rochan2018video}  & 47.5 & 56.8 \\
GLRPE\shortcite{jung2020global}  &  50.20 &  56.10 \\ 
MC-VSA\shortcite{liu2020transforming}  & 51.6  & 63.7 \\
CLIP-It\shortcite{NEURIPS2021_7503cfac}  & \textbf{54.2}  &  \textbf{66.3} \\
\midrule
Ours(SELF-VS)     & 39.80  & 58.99 \\
\bottomrule
\end{tabular}
\caption{Experimental results of F-score on SumMe and TVSum datasets. The first row reports the results from unsupervised and self-supervised methods while the second row reports the ones from supervised methods.}
\label{table:3}
\end{table}

\subsubsection{Evaluation metrics}
Similar to previous methods, we evaluate our approach by comparing video summaries to human-created summaries and reporting on F-score, which is widely used, as well as correlation-based metrics, namely Kendall's $\tau$and Spearman's $\rho$.\\These metrics provide a comprehensive evaluation of the performance of our approach and allow for a better comparison with other state-of-the-art methods.

\textbf{F-score} F-score measures the overlap between human and generated video summary. More specifically, given Precision(P) and Recall(R), F-score is measured by $F = \frac{2RP}{R + P}$ where

\begin{equation}
R=\frac{V_h \cap V_g}{V_h}, P=\frac{V_h \cap V_g}{V_g}
\end{equation}
and $V_h, V_g$ are user and generated video summaries.\\
\textbf{Correlation based metrics}
\cite{smeaton2006evaluation} proposed correlation-based evaluation to address the limits of F-score. Two ranked correlation coefficients, Kendall's $\tau$ and Spearman's $\rho$ are used as metrics between the ranking of frames based on their ground truth and predicted frame-level scores.\\
\textbf{Multiple user annotations} Both F-score and correlation metrics are calculated on multiple user annotations. The average of these metrics is taken to achieve the final result.

\subsection{Implementation details}
\subsubsection{Data pre-processing}
In order to reduce the number of redundant frames in our dataset, we downsampled all videos to two frames per second for both pre-training and fine-tuning. We used the representation of the last layer of the R3D18 model \cite{tran2018closer} that was trained on the kinetics dataset \cite{carreira2017quo} as the video representation. Following the method of previous studies \cite{zhang2016video, mahasseni2017unsupervised}, we processed each frame through GoogleNet \cite{Szegedy2015GoingDW} and used the output of the pool5 layer as the frame representation.

\subsubsection{Pre-training}
We pre-trained a 6-layer transformer encoder with a batch size of 512. We used Adam to optimize over 30 epochs with a learning rate of 1e-4 with a linear warm-up on the first 5 epochs and a cosine learning rate decay for the remaining 25 epochs. Our main model uses cross-entropy as knowledge distillation loss.
\subsubsection{Fine-tuning}
To ensure robust evaluation, the model is trained and evaluated on five different data splits for each dataset. To facilitate comparison, we used the five splits provided in \cite{zhang2016video}. We employed batch size 4 and Adam optimization with a learning rate of 1e-3, betas 0.9, 0.999, and weight decay of 1e-4. The network is optimized over 25 epochs. The encoder comprised of 6 layers with 4 heads and a model dimension of 256, resulting in 64-dimensional heads. To further enhance performance, we implemented dropout with a probability of 0.2 after applying positional encoding to features.

\begin{figure*}
    \centering
    \includegraphics[width=1 \textwidth, height= 0.4 \textheight]{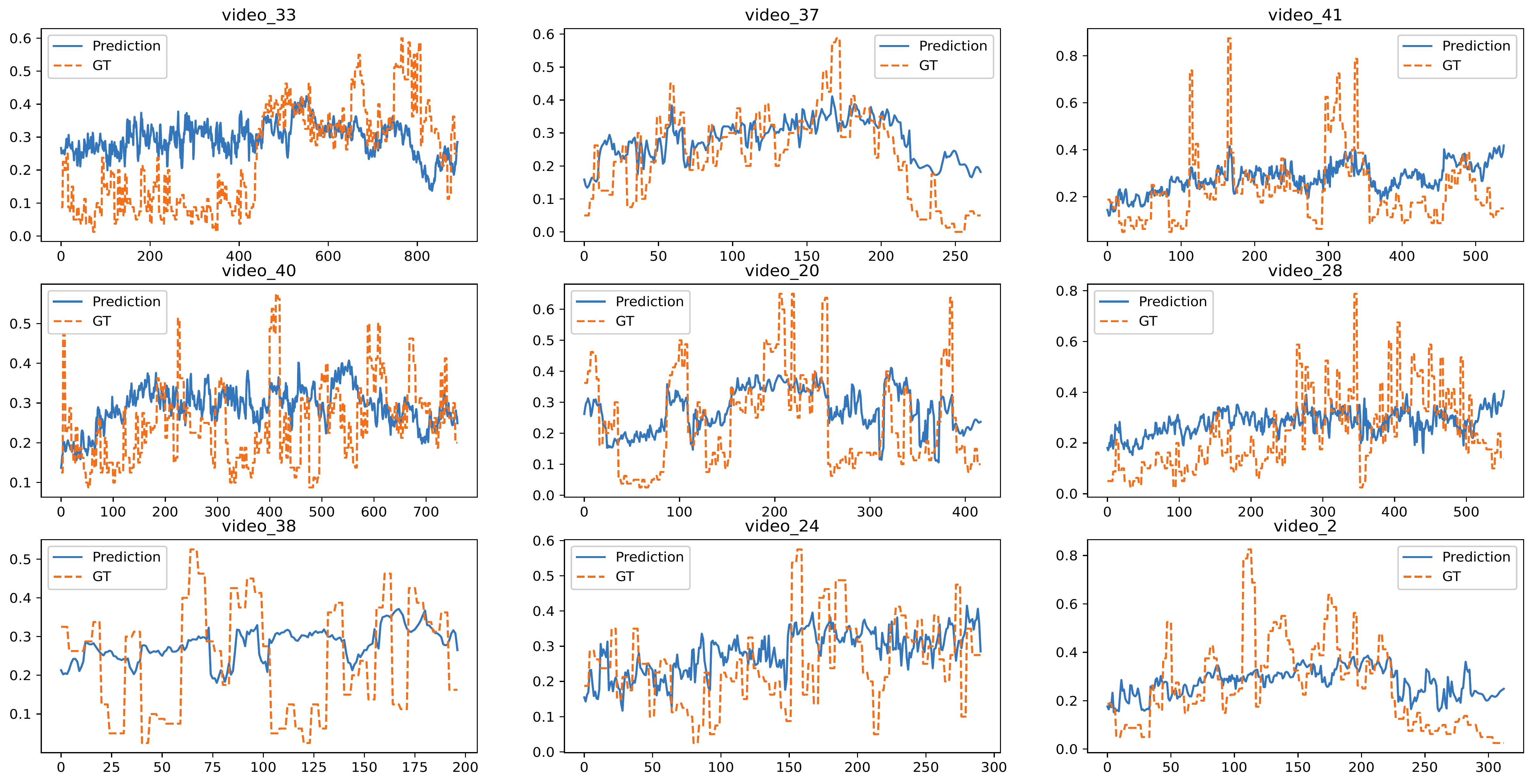}
    \caption{Frame level importance score comparison between models prediction and GT. Videos were selected from TVSum's first split validation set. Although predicted scores are not the same as GT ones, they match the ups and downs pattern of the GT graph which implicates correct frame ranking.} 
    \label{fig:3}
\end{figure*}
\subsubsection{Infrastructure details}
All experiments were conducted on the Arch Linux operating system and PyTorch\cite{NEURIPS2019_9015} version 1.12.1. We use RTX3060 paired with an intel core i5 processor and 16 gigabytes of memory.  

\begin{table}
\centering
\begin{tabular}{cccc}
\toprule
Method & Kendall's $\tau$ & Spearman's $\rho$\\
\midrule
Random & 0.000 & 0.000 \\
Human   & 0.177 & 0.204 \\
Ground Truth & 0.364  &  0.456\\
\midrule
SGAN\shortcite{Mahasseni_2017_CVPR}         & 0.024 &  0.032 \\
DR-DSN\shortcite{zhou2018deep}             & 0.020     & 0.026 \\
SumGraph\shortcite{Park2020SumGraphVS}  &       0.094  &  0.138 \\
AudViSum\shortcite{Chowdhury2021AudViSumSD}  & 0.101  &  0.146 \\
SSPVS\shortcite{haopeng2022video}  & 0.169 & 0.231 \\
\midrule
dppLSTM\shortcite{zhang2016video}           & 0.042     & 0.055 \\
HSA-RNN \shortcite{zhao2018hsa}          & 0.082 & 0.088 \\
GLRPE\shortcite{jung2020global}          & 0.070     & 0.091 \\
MC-VSA\shortcite{liu2020transforming}            & 0.116     & 0.142 \\
RSGN\shortcite{Zhao2022ReconstructiveSN}  &     0.083  & 0.090 \\
\midrule
Ours(SELF-VS)                & \textbf{0.176}     & \textbf{0.232}\\
\bottomrule
\end{tabular}
\caption{The results of Rank-based metrics on TVSum dataset. The second row reports results from some unsupervised and self-supervised methods. The third row shows the results from supervised methods.}
\label{table:2}
\end{table}

\subsection{Comparison with other methods}

In table \ref{table:2} and \ref{table:3} we compare our method to other video summarization methods on TVSum and SumMe datasets. We report published results by authors as well as our approach. The results are in terms of F-score and correlation metrics i.e. Kendall's $\tau$ and Spearman's $\rho$. 

\subsubsection{F-measure}
It is clear from Table \ref{table:3} that our approach did not achieve a high F-score. This may be due to the fact that the F-score metric does not take into account a good relative score, which is the focus of our pre-training. Additionally, as noted in \cite{otani2019rethinking}, the F-score is highly dependent on the selection of video segmentation, such as KTS segmentation.

\subsubsection{Ranked-based metrics}
On the other hand in table \ref{table:2}, our approach outperforms state of art methods by a large margin on correlation metrics which suggests its superior strength in producing relatively good scores for frames.

% end experiment

%\newcommand{\ra}[1]{\renewcommand{\arraystretch}{#1}}
%\begin{table*}
%\centering

%\ra{1.0}
%%\begin{tabular}{clccccccc} \toprule & \multicolumn{3}{c}{CE} & \multicolumn{4}{c}{MSE}\\
%\cmidrule{2-4}
%\cmidrule{6-8}
%$\lambda$ & F-score & Kendall's $\tau$ & Spearman's $\rho$ 
%&  & F-score & Kendall's $\tau$ & Spearman's $\rho$\\
%\midrule
%0.001 & 59.50 & 0.174 & 0.230 & & 59.50 & 0.174 & 0.230 \\
%0.01 & 58.81 & 0.174 & 0.230 & & 58.81 & 0.174 & 0.230 \\
%0.07 & 58.56 & 0.169 & 0.222 & & 58.56 & 0.169 & 0.222 \\
%0.3 & 58.88 & 0.168 & 0.221 & & 58.43 & 0.170 &  0.224 \\
%0.15 & 59.10 & 0.169 & 0.223 & & 59.00 & 0.172 & 0.227\\
%0.6 & 58.87 & 0.173 & 0.228 & & 59.35 & 0.173 & 0.229\\
%1.0 & 58.95 & 0.173 & 0.228 & & 58.81 & 0.170 & 0.224 \\
%2.0 & 58.61 & 0.177 & 0.233 & & 58.61 & 0.177 & 0.233 \\

%\bottomrule
%\end{tabular}
%\caption{Comparison between different $\lambda$s on TVSum. We study the performance of the %model under different regularization intensities.}
%\label{table:5}
%\end{table*}

\subsection{Ablation study}
%\subsection{Impact of components}
In this section, we analyze the impact of different components from pre-training to fine-tuning on our approach to video summarization.

Table \ref{table:4} presents the performance of our model using different schemes. As can be seen, pre-training the model has a significant impact on performance, which suggests the validity of our pre-training task.

\begin{table}
\centering
\aboverulesep=0ex % Solution part 1 of 3
   \belowrulesep=0ex % Solution part 1 of 3
%   \caption{Example records from a data frame.}
\begin{tabular}{cc|ccc}
\toprule
\rule{0pt}{1.1EM}% Solution part 2 of 3 (% is required)
Pretrain      & Loss      & F-score       &Kendall's$\tau$      &  Spearman's$\rho$ \\
% \rule{0pt}{1.1EM}
\midrule 
\rule{0pt}{1.1EM} %
\cmark  &CE      &\textbf{58.99}      &\textbf{0.176}          &\textbf{0.232}\\
%\cmark         &\xmark                      &CE      & 58.75        & 0.11         &0.13\\
\cmark  &MSE     &58.19       &0.168          &0.221\\
%\cmark         &\xmark                      &MSE     & 56.40       & 0.02         & 0.03\\
\xmark         &\_  & 58.34  & 0.103     & 0.136\\

\bottomrule
\end{tabular}
\caption{Effect of different components on SELF-VS. Models are trained based on the default settings and the results are reported on TVSum.}
\label{table:4}
\end{table}

Additionally, we experiment with different loss functions to understand the influence of these on pre-training. Results show that cross-entropy outperforms mean squared error by a small margin, indicating the robustness of our approach to the choice of loss function. As the figure \ref{fig:2} depicts our pre-training scheme aims at predicting relatively meaningful scores for frames. Compared to the random network predicted scores are successful in achieving the overall pattern of the ground truth graph.

\begin{figure}
    \centering
    \includegraphics[width=0.4 \textwidth]{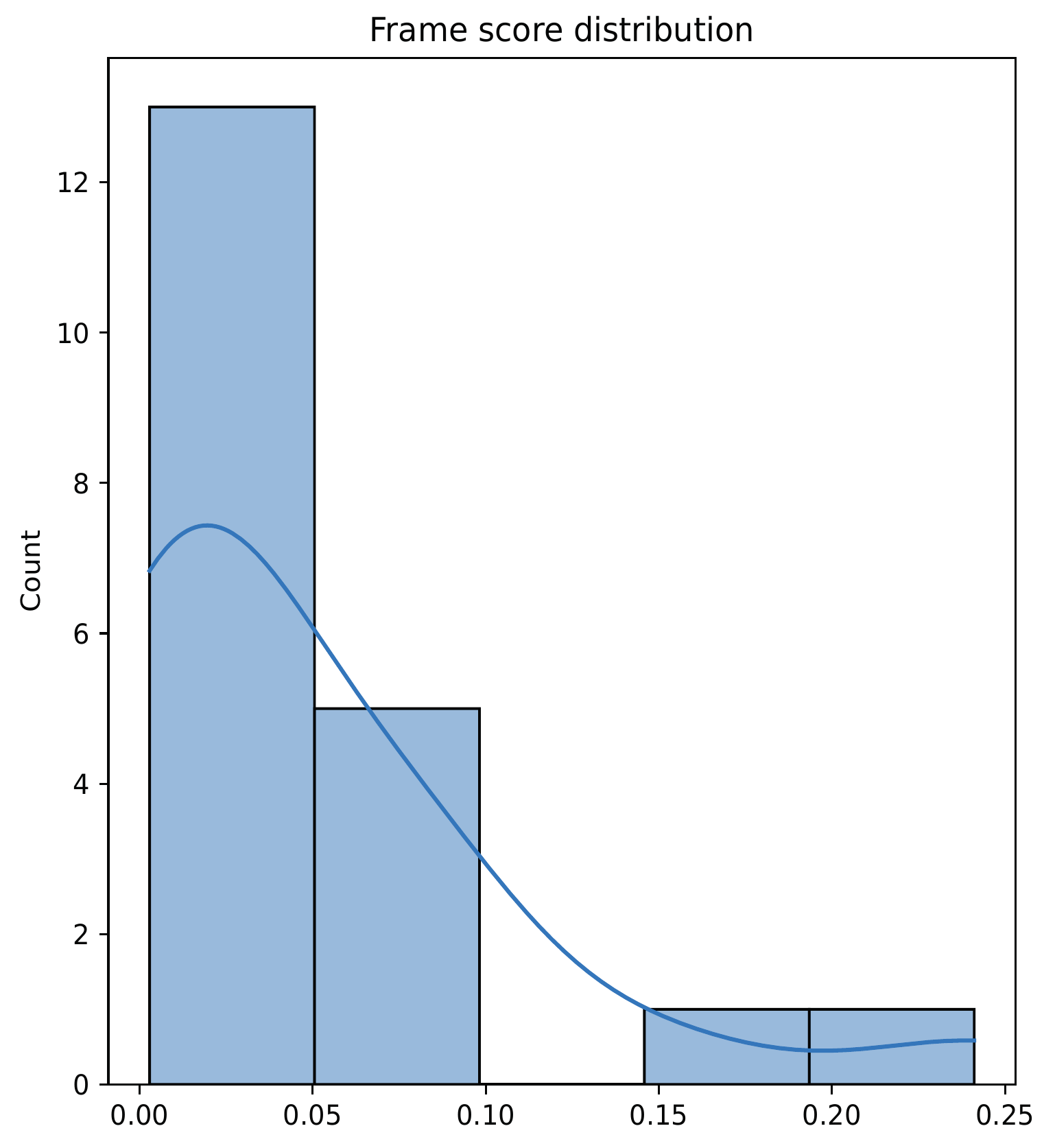}
    \caption{Frames score histogram generated by model in pre-training stage. These scores are used as weights in a weighted sum to construct encoder video representation.} 
    \label{fig:4}
\end{figure}

\subsection{Possibility of Collapse}
During pre-training, the network may try to cheat by assigning the same score to all frames or only considering a few frames to construct the video representation of the encoder. However, as shown in Figure 3 \ref{fig:3}, if the model had collapsed during pre-training, the predicted score graph would have been a relatively straight or highly unstable line, indicating that the mentioned collapse does not occur. To further validate this, we visualized the score distribution of the encoder while pre-training in Figure 4\ref{fig:4}. It is clear that neither a uniform nor a dimensional collapse affects our pre-training approach.

\subsection{Visualizing predicted score}
Figure \ref{fig:3} visualizes ground truth scores and model predicted ones from TVSum's first split's validation set after fine-tuning on the training set. Though the model is fine-tuned on the ground truth score, it does not fit the ground truth graph which explains the poor performance on F-measure. However, it assigns relatively correct scores to frames which results in matching frame ordering based on prediction and ground truth scores.

\begin{figure}
    \centering
    \includegraphics[width=0.4 \textwidth, height=0.3 \textheight]{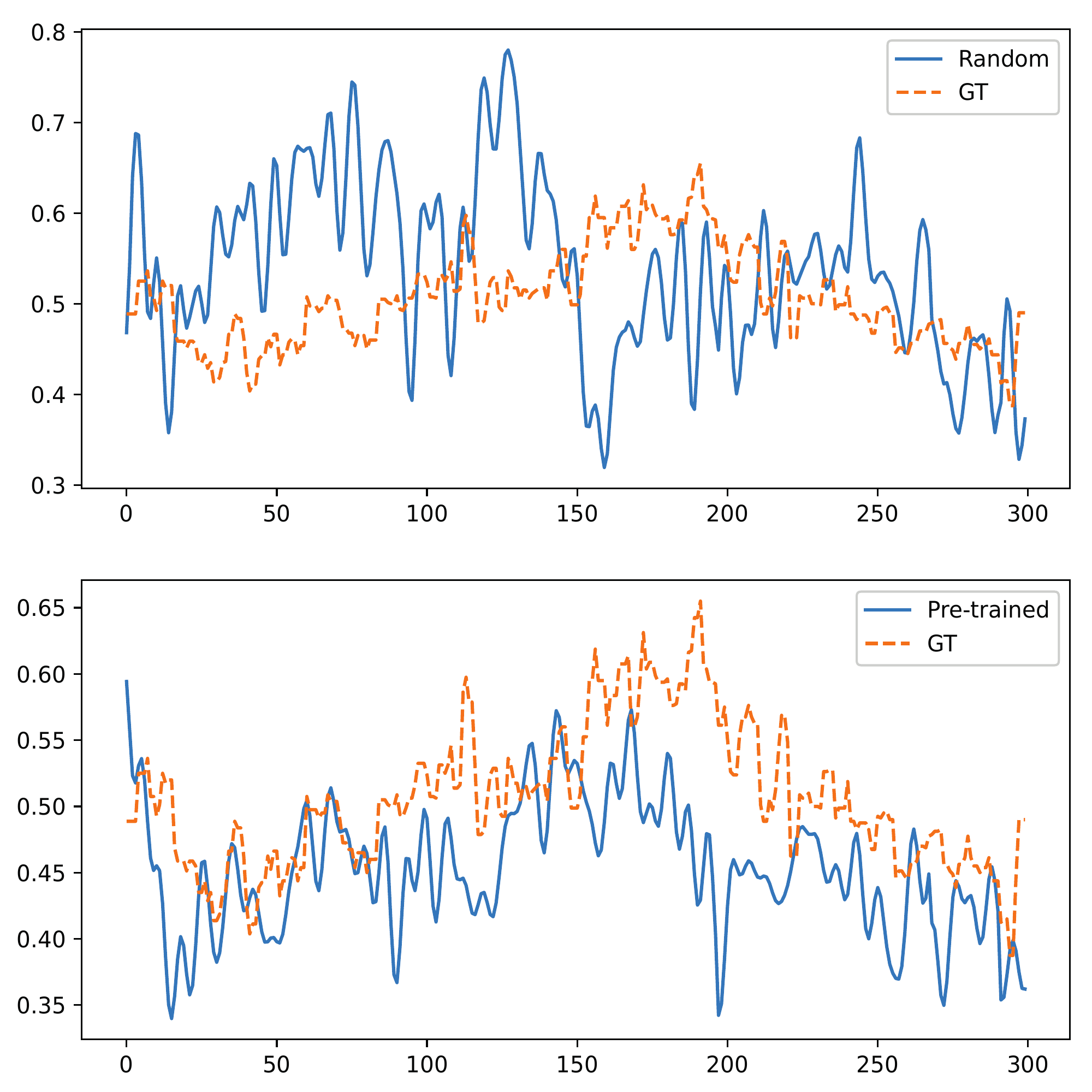}
    \caption{Comparison between randomly initialize network and pre-trained network generated importance score and ground truth averaged over first 300 frames of all videos.} 
    \label{fig:2}
\end{figure}
%\subsection{Collapse study}
%We study the effect of collapse prevention by maximizing the diversity of representation of frames. In table \ref{table:5} we compared performance of model under different regularization intensities. $\lambda$ of 0.6 results in the highest values for correlation metrics, however, it only outperforms other values by a low margin which shows the robustness of regularization to the choice of coefficient $\lambda$. 
%ends ablation

\section{Limitation}
Despite the success of our method at assigning relatively good importance scores to video frames, it lags behind when it comes to assigning the exact annotated frame scores. This is in part due to the specific choice of pre-training which does not incorporate true frame scores alongside relative scores. Exploring ways to account for this problem is left as future work.

\section{Conclusion}
In this work, we offered a novel self-supervised scheme to deal with the video summarization task. Experimental results suggest that our approach of utilizing an off-the-shelf representation of video as the self-supervision signal and pre-training in a teacher-student manner could be a promising route to learn an effective summary representation.\\ 
Although our method falls behind in F-score measurements, it achieves state-of-the-art on Rank-based metrics. We suggest that such behavior is in part due to the pre-training strategy which enforces the model to learn relatively decent frame level importance score.%. However, as mentioned in the experiment and result section F-score is highly dependent on the choice of segmentation method which suggests it may not be a reliable evaluation metric. \\
% to compare video summarization methods. \\

%The major limitation in our architecture is the memory and computational burden of self-attention. Even though self-attention is the necessary building-blocks of the model, it avoids us from experimenting on more lengthy videos owing to its quadratic scaling with sequence length. This issue could be mitigated via choosing more efficient self-attention architectures such as \cite{Kitaev2020ReformerTE} and \cite{Zaheer2020BigBT} which is left as the future work. 

% In this work we offered a method to deal with video summarization task with a novel self-supervise perspective. We argued that the lack of training data results in bad performance and over-fitting of deep neural networks. To solve this, we proposed to re purpose a 3D ResNet as a teacher to pre-train a transformer network in knowledge distillation fashion. Our results indicate soundness of our approach on correlation based metrics, however it fails to achieve high F-score. We further studied the impact of different components in the pipeline through extensive experiments.

\bibliography{aaai23}
\end{document}